\PassOptionsToPackage{table}{xcolor}

\documentclass[sn-basic,iicol]{sn-jnl}

\usepackage{amsmath,amssymb,amsfonts}
\usepackage{amsthm}
\usepackage{mathrsfs}
\usepackage{bm}

\usepackage{graphicx}
\usepackage{xcolor}
\usepackage{colortbl}
\usepackage{booktabs}
\usepackage{array}
\usepackage{multirow}
\usepackage{makecell}
\usepackage{tabularx}
\usepackage{longtable}
\usepackage{rotating}
\usepackage{overpic}
\usepackage{float}
\usepackage{wrapfig}
\usepackage{sidecap}
\usepackage{caption}
\usepackage{subcaption}

\usepackage{algorithm}
\usepackage{algorithmicx}
\usepackage{algpseudocode}
\usepackage{listings}

\usepackage[title]{appendix}
\usepackage{textcomp}
\usepackage{manyfoot}
\usepackage{pifont}
\usepackage{times}
\usepackage{microtype}
\usepackage[normalem]{ulem}
\usepackage{enumitem}
\usepackage{footmisc}
\usepackage{siunitx}
\usepackage{bbding}
\usepackage{xspace}
\usepackage{comment}
\usepackage{tcolorbox}
\usepackage{placeins}

\usepackage{tikz}
\usetikzlibrary{patterns}

\usepackage{pgfplots}
\usepgfplotslibrary{groupplots}

\usepackage{hyperref}
\usepackage[capitalize]{cleveref}

\definecolor{myblue}{HTML}{92dcec}
\definecolor{hous}{HTML}{b88b4d}
\definecolor{green}{HTML}{79c561}
\definecolor{farming}{HTML}{ded94c}
\definecolor{trans}{HTML}{b4b4a9}
\definecolor{services}{HTML}{ff362e}
\definecolor{other}{HTML}{dbd4d3}
\definecolor{industry}{HTML}{db79c0}
\definecolor{water}{HTML}{7982db}
\definecolor{techinfra}{HTML}{303355}
\definecolor{aliceblue}{rgb}{0.94, 0.97, 1.0}

\graphicspath{{./}}

\Crefname{section}{Section}{Sections}
\Crefname{table}{Table}{Tables}

\makeatletter

\newcommand{\FullRowBackground}[1]{%
  \makebox[0pt][l]{%
    \begingroup
    \setlength{\fboxsep}{0pt}%
    \colorbox{#1}{%
      \rule[-\dp\@arstrutbox]{0pt}{%
        \dimexpr\ht\@arstrutbox+\dp\@arstrutbox\relax}%
      \hspace{\textwidth}%
    }%
    \endgroup
  }%
}

\DeclareRobustCommand\onedot{\futurelet\@let@token\@onedot}
\def\@onedot{\ifx\@let@token.\else.\null\fi\xspace}

\makeatother

\begin{document}

\title[RATR: Role Assignment and Temporal Routing]{
RATR: Training-Free Material Rebinding via Role Assignment and Temporal Routing
}


\author[1]{\fnm{Jiehui} \sur{Tang}}
\email{jiehuitang@zju.edu.cn}

\author*[1]{\fnm{Yunlong} \sur{Yu}}
\email{yuyunlong@zju.edu.cn}

\author[2]{\fnm{Zhao} \sur{Wang}}
\email{zhao\_wang@zju.edu.cn}


\author[3]{\fnm{Jungong} \sur{Han}}
\email{jungonghan77@gmail.com}

\affil[1]{
    \orgdiv{College of Information Science and Electronic Engineering},
    \orgname{Zhejiang University},
    \orgaddress{\city{Hangzhou}, \country{China}}
}


\affil[2]{
    \orgname{Ningbo Innovation Center, Zhejiang University},
    \orgaddress{\city{Ningbo}, \country{China}}
}


\affil[3]{
    \orgdiv{Department of Automation},
    \orgname{Tsinghua University},
    \orgaddress{\city{Beijing}, \country{China}}
}

\abstract{Material rebinding aims to transfer the material appearance from a reference image while preserving the target object's geometry and identity. Existing methods typically struggle to balance material fidelity with structural preservation due to the coupled influence of structural and appearance cues in their visual conditions. We identify role entanglement in the target and reference conditions as a contributing factor: the target condition carries both geometry and its original appearance while the reference condition contains material cues alongside object-specific structure. To address this issue, we propose RATR, a training-free material rebinding method built on a diffusion-based image editor that assigns complementary roles to the two conditions and coordinates their influence during denoising. To make the target condition primarily convey structure, we introduce an albedo-resolved surface release mechanism, which suppresses the target’s original surface appearance while retaining geometric cues, yielding a structure-oriented condition for material transfer. On the reference side, the designed material evidence isolation mechanism emphasizes material cues while reducing interference from object-specific structure, yielding a complementary material-oriented condition. Surface–material conditional direction routing then dynamically adjusts the residual contributions of these two conditions at different denoising stages, balancing structural preservation and material transfer without modifying the editor’s parameters. Across diverse target–material pairs, RATR delivers superior results on all five reported target-side preservation metrics and leading reference-side material fidelity under our evaluation protocol. The project repository is available at \protect\url{https://github.com/TangJieH/RATR}.

}

\keywords{
Material rebinding,
Role assignment,
Temporal routing,
Training-free image editing,
Material evidence isolation
}

\maketitle


\section{Introduction}
\label{sec:introduction}

\begin{figure*}[t]
\centering
\includegraphics[width=\textwidth]{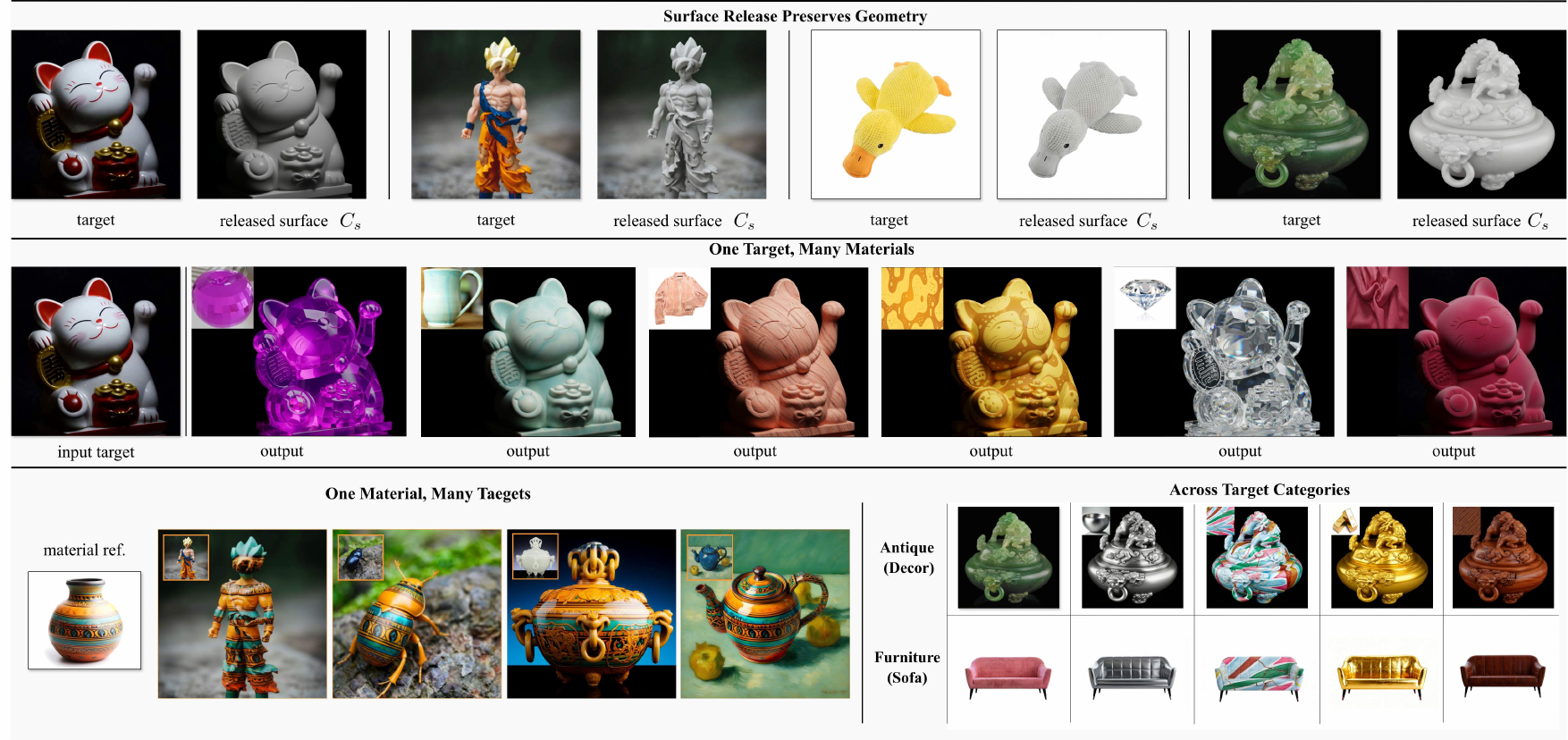}
\caption{
Qualitative demonstration of RATR. 
Top: surface release produces albedo-resolved surface
boundaries $C_s$ that preserve target geometry while
suppressing the original target appearance. 
Middle: a fixed target can be rebound to diverse reference materials. 
Bottom: a single material can transfer across different targets}
\label{fig:teaser}
\end{figure*}

Given a target image and a reference material exemplar, material rebinding aims to change how the target surface appears while preserving what the target object is. This requires transferring material-specific appearance without altering the target's geometry, identity, or object-specific details. The difficulty lies in the visual evidence itself: the target image contains not only the structure to be preserved but also the appearance to be replaced, whereas the reference image contains not only the material to be transferred but also the object and scene in which that material is observed. Successful rebinding therefore requires selective use of both images.

Recent diffusion-based material transfer methods introduce explicit material exemplars together with auxiliary geometric, lighting, depth, normal, or token-level conditions \citep{cheng2024zest,garifullin2025materialfusion,lopes2025matswap,huang2025mate,huang2026dealmate}. Although these designs improve material fidelity, stronger material guidance can also distort target geometry, over-smooth surface details, or introduce reference-object content. Large-scale image editors such as Qwen-Image-Edit provide an alternative through powerful reconstruction priors that help preserve object identity and geometry~\citep{wu2025qwenimage,qwen2025imageedit2511}. However, preserving the target too strongly can retain its original surface appearance and limit material transfer. These outcomes raise a central question: why does strengthening the evidence for one objective often undermine the other?

We examine this question through the information carried by the two visual conditions. The target condition jointly encodes desired structure and the original target appearance; the reference condition jointly encodes transferable material cues and undesired reference-object content. Increasing the strength of either condition does not selectively amplify the information needed for rebinding. Stronger target conditioning can preserve geometry while reinforcing the original material, whereas stronger reference conditioning can improve material appearance while increasing reference-content interference. Thus, structural distortion and incomplete material transfer can be understood as two manifestations of the same underlying difficulty: useful and interfering information are strengthened together.

We refer to this difficulty as \emph{role entanglement}. Unlike a conflict between otherwise well-specified conditions, role entanglement occurs within each condition, whose visual content extends beyond its intended responsibility. This distinction matters because adjusting the relative strengths of the raw target and reference conditions leaves their internal mixtures intact. Material rebinding therefore requires more than deciding which image should exert greater influence: it requires specifying which information each image should contribute. Moreover, these contributions need not remain fixed throughout diffusion synthesis. Structural evidence can establish the target geometry in earlier denoising stages, while material evidence can guide surface appearance in later stages. This motivates treating condition content and its temporal allocation as two complementary design problems.

Accordingly, we reformulate material rebinding as a \emph{role assignment and temporal routing} problem. Role assignment constructs a target condition focused on structure and identity and a reference condition focused on transferable material appearance. Temporal routing then coordinates their contributions over the denoising process. This formulation addresses two limitations of raw two-image conditioning: each input carries information outside its intended role, and a fixed allocation of condition influence does not account for the different requirements of structural preservation and material synthesis.

Based on this formulation, we propose \textbf{RATR}, a training-free material rebinding method built on a frozen two-image diffusion editor. On the target side, \textbf{Albedo-Resolved Surface Release} constructs albedo-resolved surface boundaries that retain geometric cues while suppressing the original surface appearance. On the reference side, lightweight \textbf{Material Evidence Isolation} reduces object and scene interference to provide more focused material evidence. These operations reparameterize the inputs toward complementary structural and material roles. \textbf{Surface-Material Conditional Direction Routing (SM-CDR)} then allocates their centered VAE condition residuals over denoising time, emphasizing structural evidence earlier and material evidence later. In this way, RATR addresses both what each condition carries and when it contributes, without modifying or retraining the underlying editor.

Experiments across diverse target objects and material exemplars show consistent improvements in target-side structural preservation together with strong reference-side material fidelity. The qualitative results in Fig.~\ref{fig:teaser} illustrate both directions of generalization: a fixed target can be rebound to different materials, and a single material exemplar can be transferred across different targets. These results support the effectiveness of coordinating role-specific conditions and their temporal contributions for material rebinding.

Our contributions are summarized as follows:
\begin{itemize}
\item We identify \textbf{role entanglement} within target and reference conditions as a contributing factor to the trade-off between structural preservation and material fidelity, and reformulate material rebinding as a \textbf{role assignment and temporal routing} problem.


\item We propose \textbf{RATR}, a training-free framework that makes the target image primarily guide structure and the reference image primarily guide material appearance, and coordinates their influence across generation stages to reduce source-appearance retention and reference-content interference.

\item We evaluate RATR across diverse target--material pairs, demonstrating consistent gains in target-side structural preservation alongside strong reference-side material fidelity.

\end{itemize}

\section{Related Work}
\label{sec:related_work}

\subsection{Diffusion-Based Image Editing and Reference Conditioning}
Diffusion models have become a foundation for image synthesis and editing, with latent diffusion, high-resolution extensions, and transformer backbones expanding the quality and flexibility of generation \citep{ho2020ddpm,rombach2022ldm,podell2024sdxl,peebles2023dit}. Image-editing methods such as SDEdit and Blended Diffusion modify images through guided denoising and localized synthesis \citep{meng2022sdedit,avrahami2022blended}. Subsequent approaches improve editability and content preservation through cross-attention control, image inversion, instruction following, mask prediction, feature reuse, or mutual self-attention \citep{hertz2023prompttoprompt,mokady2023nulltext,brooks2023instructpix2pix,couairon2023diffedit,tumanyan2023plugandplay,cao2023masactrl}. Together, these works establish strong image-editing priors, but do not by themselves specify how the geometric identity of a target and the appearance of a separate material exemplar should be assigned different conditioning roles.

Controllable generation provides several mechanisms for introducing visual evidence. ControlNet and T2I-Adapter inject spatial cues such as edges, depth, and layout, whereas Composer and Uni-ControlNet investigate the composition of heterogeneous controls \citep{zhang2023controlnet,mou2024t2iadapter,huang2023composer,zhao2023unicontrolnet}. FreeControl further explores training-free spatial and appearance guidance \citep{mo2024freecontrol}. Image-prompt and exemplar-based approaches, including IP-Adapter, Paint by Example, AnyDoor, PAIR Diffusion, and BLIP-Diffusion, transfer different aspects of reference-image information for image generation or editing \citep{ye2023ipadapter,yang2023paintbyexample,chen2024anydoor,goel2024pair,li2023blipdiffusion}. Recent general-purpose editors such as Qwen-Image-Edit-2511 also support multi-image editing and material replacement \citep{wu2025qwenimage,qwen2025imageedit2511}. Nevertheless, the presence of two visual inputs does not guarantee that their contributions are restricted to the intended task: a target may retain its previous color or texture, while an exemplar can introduce unwanted object or scene information. This motivates an explicit, material-oriented organization of visual conditions within a frozen editor, rather than an assumption that generic reference conditioning already separates geometry from transferable appearance.

\subsection{Material Appearance Editing and Material Rebinding}
Material appearance has traditionally been studied through intrinsic-image decomposition, reflectance and illumination estimation, and physically based rendering \citep{bell2014intrinsic,barron2015sirfs,pharr2016pbr}. Rendering-aware material editing and single-image SVBRDF capture recover or manipulate material-related quantities, while earlier example-based approaches transfer visual attributes through image correspondences \citep{liu2017materialediting,deschaintre2018svbrdf,liao2017deepimageanalogy}. Generative material representations provide further alternatives: MaterialGAN learns an SVBRDF prior, MatFormer models procedural material structures, and MatFusion applies diffusion models to reflectance capture \citep{guo2020materialgan,guerrero2022matformer,sartor2023matfusion}. These directions provide explicit or learned material representations, but their primary objectives differ from directly assigning the appearance of an arbitrary material reference to an existing target object while maintaining its image-space structure.

Diffusion-based methods have increasingly addressed material editing without requiring an explicit material reconstruction as the final output. Alchemist learns parametric control over material properties, including roughness, metallicity, albedo, and transparency; PhyS-EdiT studies physics-aware semantic editing; and Material Palette extracts reusable material representations from a single image \citep{sharma2024alchemist,cai2025physedit,lopes2024materialpalette}. More directly, ZeST performs zero-shot material transfer using exemplar appearance together with depth and grayscale-derived target cues \citep{cheng2024zest}. MaterialFusion combines diffusion-based transfer with region constraints and a controllable transfer strength, while MatSwap learns light-aware material transfer from supervised data \citep{garifullin2025materialfusion,lopes2025matswap}. MARBLE represents material attributes and their combinations in CLIP space \citep{radford2021clip,cheng2025marble}. Related tasks include CoatFusion's application of controllable material coatings and LineArt's reference-appearance transfer to detailed design drawings \citep{levy2025coatfusion,wang2025lineart}; these settings are related to, but not identical to, material replacement on natural-image targets.

Recent diffusion-transformer methods further clarify the importance of visual-condition organization. MaTe fuses material, depth, and illumination images as tokens in a shared transformer space, enabling zero-shot material transfer without additional reference encoders or task-specific fine-tuning \citep{huang2025mate}. Its extension, DealMaTe, introduces depth-, normal-, and illumination-aware control through multi-dimensional Shader LoRA and specialized attention mechanisms \citep{huang2026dealmate}. Therefore, training-free transfer and a frozen backbone are not, by themselves, novel properties of material rebinding. RATR addresses a complementary setting: instead of adding learned condition-specific branches or constructing an explicit multi-control interface, it reorganizes the two image inputs of an existing frozen editor into surface-oriented and material-oriented roles. It then routes the resulting visual-condition contributions over denoising time. The focus is thus on the roles and timing of available conditioning signals, not on learning a new material representation or control adapter.

\subsection{Selective Conditioning and Denoising-Time Control}
A closely related challenge is preventing a reference image from transferring unintended content together with desired appearance. StyleAligned uses shared attention to maintain visual style consistency, and InstantStyle seeks to reduce content leakage through content subtraction and selective style-feature injection \citep{hertz2024stylealigned,wang2024instantstyle}. DEADiff explicitly disentangles style and semantic representations, whereas CSGO learns separate content and style pathways using supervised composition data \citep{qi2024deadiff,xing2025csgo}. StyleSSP modifies the sampling initialization to improve content-layout preservation, and Attention Distillation transfers visual characteristics through an attention-based optimization objective \citep{xu2025stylessp,zhou2025attentiondistillation}. Conditional Balance selects sensitive attention layers to manage competing style and content conditions \citep{cohen2025conditionalbalance}. These studies identify feature leakage and interference as important problems, but style and content separation does not directly establish a physically meaningful decomposition of material, geometry, and illumination. RATR accordingly treats its surface and material boundaries as task-oriented visual conditions, not as exact intrinsic-image factors.

Conditioning can also have different effects at different stages of denoising. Prompt-to-Prompt manipulates cross-attention to maintain edit consistency \citep{hertz2023prompttoprompt}; ProSpect associates image attributes with textual embeddings at different generation stages \citep{zhang2023prospect}. CADS gradually changes the noise applied to conditioning signals to control the diversity--fidelity trade-off, and StyleStudio uses early-stage guidance to stabilize spatial layouts during style transfer \citep{sadat2024cads,lei2025stylestudio}. Temporal Concept Dynamics further shows that the success of concept interventions varies with the denoising step, model, and concept \citep{gorgun2026temporal}. Collectively, these works motivate investigating \emph{when} a condition should influence generation; they do not imply a universal schedule in which geometry always forms before appearance. RATR instead evaluates a task-specific temporal routing policy for its two role-reparameterized visual conditions, leaving the pretrained editor weights and sampling scheduler unchanged.

\begin{figure*}[t]
\centering
\includegraphics[width=\linewidth]{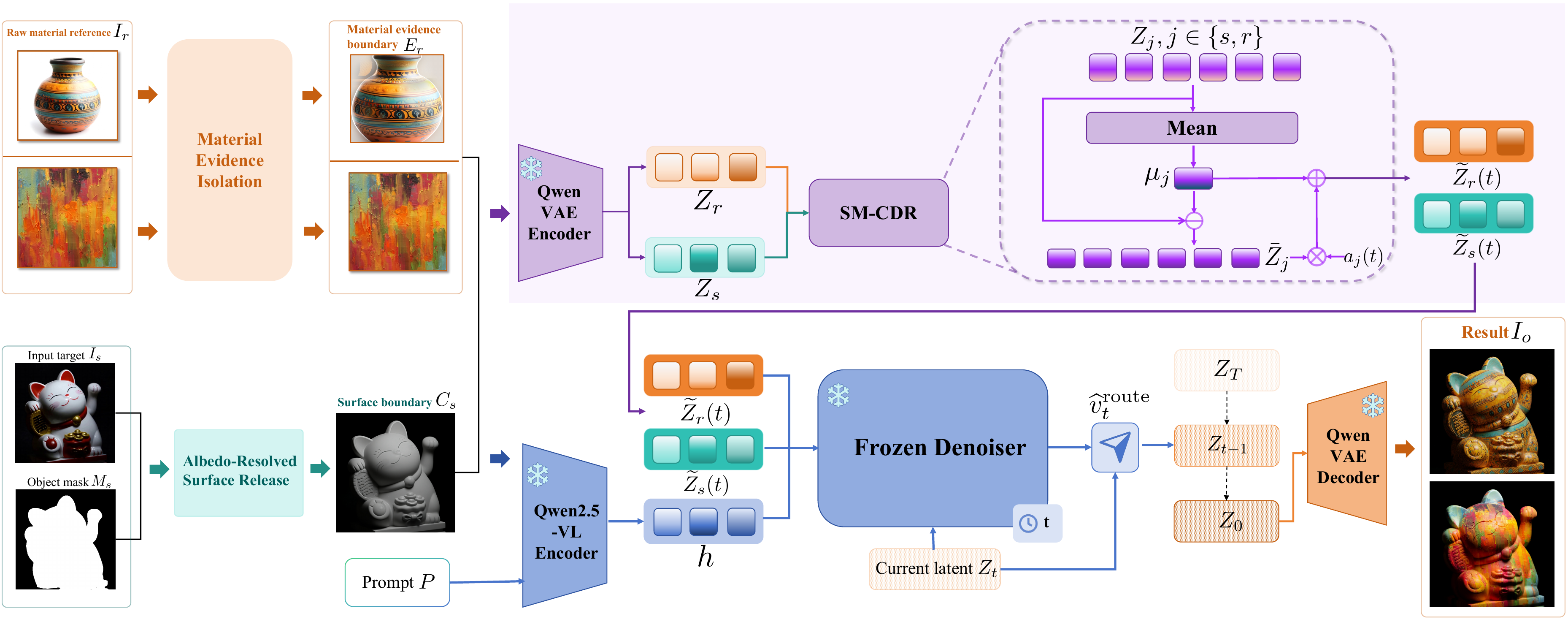}
\caption{Overview of RATR. The target and material-reference images are reparameterized into complementary visual inputs: a target-derived surface boundary $C_s$ that suppresses inherited appearance while retaining structural evidence, and a reference-derived material evidence boundary $E_r$. Surface-Material Conditional Direction Routing (SM-CDR) adjusts the contributions of their visual-condition residuals throughout denoising, emphasizing structural evidence and material evidence at different stages under a task-specific schedule. }
\label{fig:framework}
\end{figure*}


\section{Method}
\label{sec:method}

Fig.~\ref{fig:framework} illustrates RATR. We address material rebinding through two operations: role assignment constructs a target condition that retains structure while weakening its original appearance and a reference condition that emphasizes transferable material cues; temporal routing then coordinates their contributions during denoising. Both operations use a frozen two-image editor without parameter updates.

\subsection{Problem Formulation and Overview}
\label{sec:problem_formulation}

Let $I_s\in[0,1]^{H\times W\times3}$ denote the target image and $I_r$ the reference material exemplar. Given a material-rebinding instruction $P$, the objective is to synthesize an image $I_o$ that exhibits the reference material while preserving the target object's geometry, identity, and fine structural details. We use a frozen two-image editor $Q_\theta$ with fixed parameters $\theta$. The target foreground mask $M_s$ is estimated once using the U2-Net segmenter in rembg; a reference-side material support $\widehat M_r$ is obtained independently.

Under raw two-image conditioning, the editor receives $(P,I_s,I_r)$. The target image carries both the structure to preserve and the appearance to replace, while the reference contains transferable material cues together with object- and scene-specific content. Increasing either condition's influence may therefore strengthen useful and interfering information together. We refer to this within-condition mixture as \emph{role entanglement}.

RATR replaces the raw visual inputs with complementary conditions before editing:
\begin{equation}
\begin{aligned}
C_s &= \mathcal{R}_{\mathrm{surf}}(I_s,M_s),\\
E_r &= \mathcal{C}_{\mathrm{mat}}(I_r,\widehat M_r),\\
\widetilde I_o &= Q_\theta(C_s,E_r,P;\Gamma),
\end{aligned}
\label{eq:aris_overview}
\end{equation}
where $C_s$ is the target-side surface condition, $E_r$ is the reference-side material evidence condition, $\Gamma$ is the denoising-time routing policy, and $\widetilde I_o$ is the editor's decoded result before foreground compositing. We call $C_s$ and $E_r$ \emph{boundaries} to emphasize their conditioning roles, not because they are binary edge maps. Their construction favors complementary information without claiming an exact decomposition of geometry and appearance.

The frozen editor encodes the two images through both a vision-language (VL) semantic path and a VAE appearance path. The semantic condition is
\begin{equation}
h=\Phi_{\mathrm{VL}}(P,C_s,E_r),
\label{eq:semantic_condition}
\end{equation}
while the VAE image-conditioned segments are
\begin{equation}
\begin{aligned}
&Z_s=Z(C_s),\qquad Z_r=Z(E_r),\\
&Z(I)=\mathrm{Pack}\big(E_{\mathrm{vae}}(I)\big),
\end{aligned}
\label{eq:condition_segments}
\end{equation}
where $E_{\mathrm{vae}}$ is the frozen VAE encoder and $\mathrm{Pack}(\cdot)$ arranges its output into image-condition tokens. RATR constructs $C_s$ and $E_r$ in image space, then routes only the centered residuals of $Z_s$ and $Z_r$. For each target--reference pair, $h$ is computed from the reparameterized inputs and held fixed during routing. Neither the VL representation nor the editor weights are modified at individual denoising steps.

At the level of intended roles, the raw condition segments can be described schematically as
\begin{equation}
\begin{aligned}
Z(I_s)&\approx Z_G+Z_A+\epsilon_s,\\
Z(I_r)&\approx Z_{\mathrm{mat}}+Z_{\mathrm{ref}}+\epsilon_r,
\end{aligned}
\label{eq:role_mixture}
\end{equation}
where $Z_G$ refers to target structure and identity, $Z_A$ to the original target appearance, $Z_{\mathrm{mat}}$ to transferable material cues, and $Z_{\mathrm{ref}}$ to reference-object and scene content. These are conceptual components, not identifiable latent factors. Role assignment seeks to weaken $Z_A$ and $Z_{\mathrm{ref}}$ without eliminating the information needed for rebinding.

\subsection{Role Assignment through Visual Condition Construction}
\label{sec:role_assignment}

Role assignment treats the two inputs differently because they contain different sources of interference. The target needs to retain geometry without anchoring the old material; the reference needs to retain material appearance without imposing its own object or scene. We construct these conditions independently so that a released surface can be reused with different materials and a material evidence boundary can be reused with different targets. We refer to this task-oriented separation of visual
conditions as role decoupling.

\subsubsection{Albedo-Resolved Surface Release}
\label{sec:surface_release}

Albedo-Resolved Surface Release constructs $C_s$ using a reference-free release trajectory followed by luminance-only structural restoration. Rather than attempting intrinsic image decomposition, it selects a surface that weakens the original chromatic appearance while preserving the target's geometry and visible relief.


\textbf{Reference-free candidate selection.}
Without introducing $I_r$, we obtain predicted-clean hypotheses $\widehat{x}_0(t)$ from a release trajectory at a fixed set of early steps $\mathcal{T}_{\mathrm{rel}}$. For \emph{each target image}, the release step is selected by
\begin{equation}
t^\star
=
\arg\min_{t\in\mathcal{T}_{\mathrm{rel}}}
\mathcal{E}_{\mathrm{rel}}
\big(\widehat{x}_0(t);I_s,M_s\big),
\label{eq:release_selection}
\end{equation}
where $\widehat{x}_0(t)$ is the clean-image estimate at the corresponding denoising step. The selection energy combines three fixed criteria:
\begin{equation}
\begin{aligned}
\mathcal{E}_{\mathrm{rel}}(x;I_s,M_s)
={}&\lambda_{\mathrm{alb}}\mathcal{E}_{\mathrm{alb}}(x)
+\lambda_{\mathrm{geo}}\mathcal{E}_{\mathrm{geo}}(x)\\
&+\lambda_{\mathrm{sup}}\mathcal{E}_{\mathrm{sup}}(x).
\end{aligned}
\label{eq:release_energy}
\end{equation}

The weights and candidate steps are shared across samples; only the minimizing step $t^\star$ depends on the input target. Let $Y(\cdot)$ and $\mathrm{CbCr}(\cdot)$ denote luminance and chroma in YCbCr space, $c_0$ neutral chroma, and $\lvert M_s\rvert$ the foreground area. The chroma criterion is
\begin{equation}
\mathcal{E}_{\mathrm{alb}}(x)
=
\frac{\left\lVert M_s\odot\big(\mathrm{CbCr}(x)-c_0\big)\right\rVert_1}{\lvert M_s\rvert+\varepsilon}.
\label{eq:albedo_energy}
\end{equation}

It discourages retention of the target's original color without penalizing luminance, which also carries shape and relief cues. It is a chroma-based selection criterion, not an estimate of physical albedo.

Let $\mathcal B(\cdot)$ be the gradient magnitude computed with a $3\times3$ Sobel operator. The geometry criterion compares luminance boundaries:
\begin{equation}
\mathcal{E}_{\mathrm{geo}}(x)
=
\frac{\left\lVert M_s\odot\left[\mathcal B(Y(x))-\mathcal B(Y(I_s))\right]\right\rVert_1}{\lvert M_s\rvert+\varepsilon}.
\label{eq:geometry_energy}
\end{equation}

It favors candidates that preserve the target silhouette and local boundary detail. Finally, with $\mathcal M(\cdot)$ denoting the fixed foreground estimator used to obtain $M_s$, the support criterion is
\begin{equation}
\mathcal{E}_{\mathrm{sup}}(x)
=1-\mathrm{IoU}\big(\mathcal M(x),M_s\big).
\label{eq:support_energy}
\end{equation}

This term penalizes foreground erosion, expansion, or displacement. Together, the three criteria select a sample-specific release hypothesis that weakens old color while retaining boundary and support evidence. We denote this hypothesis by $C_s^0=\widehat{x}_0(t^\star)$.


\textbf{Luminance residual restoration.}
The selected hypothesis can lose fine structural detail. Blending back the RGB target would also restore the original appearance, so we recover only a luminance-side high-frequency residual. With a fixed high-pass operator $\mathrm{HP}(\cdot)$ and a foreground-aware structural support map $W_s$, we define
\begin{equation}
\Delta_{\mathrm{str}}
=
W_s\odot\left[\mathrm{HP}(Y(I_s))-\mathrm{HP}(Y(C_s^0))\right].
\label{eq:structure_residual}
\end{equation}

The restored luminance and final surface condition are

\begin{equation}
Y_s^{\mathrm{rel}}
=Y(C_s^0)+\lambda_{\mathrm{str}}\Delta_{\mathrm{str}},
\label{eq:released_luminance}
\end{equation}
\begin{equation}
C_s
=\mathrm{Clip}\!\left[\mathrm{Merge}_{\mathrm{YCbCr}}\!\left(Y_s^{\mathrm{rel}},\mathrm{CbCr}(C_s^0)\right)\right].
\label{eq:surface_boundary_final}
\end{equation}

Thus, structural residuals are restored through luminance, whereas chroma remains inherited from the selected release hypothesis. The resulting $C_s$ is intended as a geometry-oriented visual condition; it does not recover intrinsic geometry or guarantee removal of every appearance cue.

\subsubsection{Material Evidence Isolation}
\label{sec:material_evidence}

Material Evidence Isolation constructs $E_r$ from the reference image without accessing the target. Reference exemplars may contain background layout, object contours, or cast shadows alongside their material cues. We therefore use a lightweight, mode-aware image-space operation that reduces obvious non-material interference while retaining visible material appearance:
\begin{equation}
E_r=\mathcal{C}_{\mathrm{mat}}(I_r,\widehat M_r),
\label{eq:material_evidence}
\end{equation}
where $\widehat M_r$ is the effective material support and $R$ denotes the editor-side evidence resolution. When the reference already provides a full-field material observation, the operation is a direct resize:
\begin{equation}
E_r=\mathrm{Resize}_{R}(I_r).
\label{eq:full_field_material_canvas}
\end{equation}

For object-centric references, $\widehat M_r$ is estimated automatically using the same U2-Net foreground extractor as on the target side. The material-bearing region is retained, while the complementary region is completed with low-frequency context derived from $I_r$. The operator fits the effective support to the material evidence boundary, softly reinforces representative local material cues, and applies weak smoothing and range normalization. References whose material support already occupies most of the image are processed in full-field mode; no pair-specific manual mask is required.

This process reduces conspicuous reference-object and scene cues without claiming to separate material from geometry exactly. In particular, $E_r$ remains a continuous image condition, not a texture atlas or an estimate of SVBRDF, lighting, or surface normals. Together with $C_s$, it provides more role-consistent visual evidence; temporal routing determines how that evidence influences the frozen editor.

\subsection{Surface-Material Conditional Direction Routing}
\label{sec:smcdr}

After constructing $C_s$ and $E_r$, RATR supplies them to the editor in the fixed order $(C_s,E_r)$. Their VAE condition segments $Z_s\in\mathbb R^{N_s\times d}$ and $Z_r\in\mathbb R^{N_r\times d}$ are routed during denoising, while the semantic condition $h$ in Eq.~\ref{eq:semantic_condition} remains fixed. SM-CDR changes neither the generation latent directly nor the prompt, VL condition, model parameters, positional encodings, or scheduler.

\subsubsection{Centered Visual-Condition Residual Modulation}

For $j\in\{s,r\}$, we separate each image-condition segment into its token mean and centered residual:
\begin{equation}
\mu_j=\frac{1}{N_j}\sum_{i=1}^{N_j}Z_j^{(i)},
\qquad
\bar Z_j=Z_j-\mu_j.
\label{eq:segment_mean_residual}
\end{equation}

SM-CDR scales the centered residual, leaving the segment mean unchanged:
\begin{equation}
\widetilde Z_j(t)
=\mu_j+\alpha_j(t)\bar Z_j.
\label{eq:direction_routing_modulation}
\end{equation}

The perturbation relative to the original segment is
\begin{equation}
\delta Z_j(t)
=\widetilde Z_j(t)-Z_j
=\big(\alpha_j(t)-1\big)\bar Z_j.
\label{eq:latent_perturbation}
\end{equation}

Unlike uniform scaling of $Z_j$, this operation does not change the segment mean. It does not, however, preserve the full latent distribution or isolate individual physical attributes. Its purpose is to regulate the variation carried by each role-specific condition.

\subsubsection{Denoising-Time Routing Policy}

We use a fixed three-stage policy to change the relative influence of surface and material evidence. With zero-based denoising-step index $k$, a total of $T$ steps, and normalized progress $u_t=k/(T-1)$, two boundaries $u_0<u_1$ define
\begin{equation}
\tau(t)=
\begin{cases}
\mathrm{early}, & u_t<u_0,\\
\mathrm{mid}, & u_0\leq u_t<u_1,\\
\mathrm{late}, & u_t\geq u_1.
\end{cases}
\label{eq:routing_segments}
\end{equation}

Each stage $\tau$ has a pair of gains $(\alpha_s^\tau,\alpha_r^\tau)$, giving
\begin{equation}
\alpha_s(t)=\alpha_s^{\tau(t)},
\qquad
\alpha_r(t)=\alpha_r^{\tau(t)}.
\label{eq:routing_schedule}
\end{equation}

The complete policy $\Gamma$ consists of the two transition boundaries and the three surface/material gain pairs. For reporting its relative bias, we define the material-routing share
\begin{equation}
p_\tau=\frac{\alpha_r^\tau}{\alpha_s^\tau+\alpha_r^\tau},
~~
\mathbf p_\Gamma
=\{p_{\mathrm{early}},p_{\mathrm{mid}},p_{\mathrm{late}}\}.
\label{eq:material_routing_share}
\end{equation}

Smaller $p_\tau$ indicates a more surface-oriented stage and larger $p_\tau$ a more material-oriented stage. The share summarizes a policy but does not determine it uniquely: the absolute gains are also part of $\Gamma$. Our schedule emphasizes structural support earlier and material appearance later. This is a task-specific allocation, not an assumption that diffusion models universally separate geometry and material at fixed times.

\subsection{Local Interpretation of Routed Conditioning}
\label{sec:local_interpretation}

To clarify what SM-CDR changes, let $z_t$ be the current generation latent and $v_\theta$ the frozen denoiser's conditional prediction direction. At the same $z_t$ and timestep $t$, standard and routed conditioning give
\begin{equation}
\widehat v_t^{\mathrm{std}}
=v_\theta(z_t,t;h,Z_s,Z_r),
\label{eq:standard_velocity}
\end{equation}
\begin{equation}
\widehat v_t^{\mathrm{route}}
=v_\theta\big(z_t,t;h,\widetilde Z_s(t),\widetilde Z_r(t)\big).
\label{eq:routed_velocity}
\end{equation}

The unchanged scheduler advances the sample using $\widehat v_t^{\mathrm{route}}$. For small perturbations $\delta Z_s(t)$ and $\delta Z_r(t)$, a first-order expansion around the static role-reparameterized conditions gives
\begin{equation}
\begin{aligned}
\Delta v_t
&\equiv\widehat v_t^{\mathrm{route}}-\widehat v_t^{\mathrm{std}}\\
&\approx\big(\alpha_s(t)-1\big)J_s(t)\bar Z_s\\
&+\big(\alpha_r(t)-1\big)J_r(t)\bar Z_r,
\end{aligned}
\label{eq:velocity_decomposition}
\end{equation}
where $J_s(t)$ and $J_r(t)$ are the local Jacobian maps with respect to the two visual condition segments, evaluated with $h$ and $z_t$ fixed. This expression explains how the gains locally change the predicted direction through the two condition residuals. It does not imply that the Jacobians are orthogonal, that material and geometry are identifiable in VAE space, or that the resulting denoising trajectory is globally optimal.

The dependence on role assignment is equally important. Before reparameterization, the target residual also carries the old appearance, and the reference residual may carry its original object and scene. Scaling those raw residuals can strengthen the unwanted cues along with the desired ones. By constructing $C_s$ and $E_r$ first, RATR reduces these sources of interference before applying the time-dependent gains. The actual extent of interference reduction is assessed empirically, rather than inferred from the local expansion.

\begin{figure*}[!tp]
\centering
 \includegraphics[width=1\textwidth]{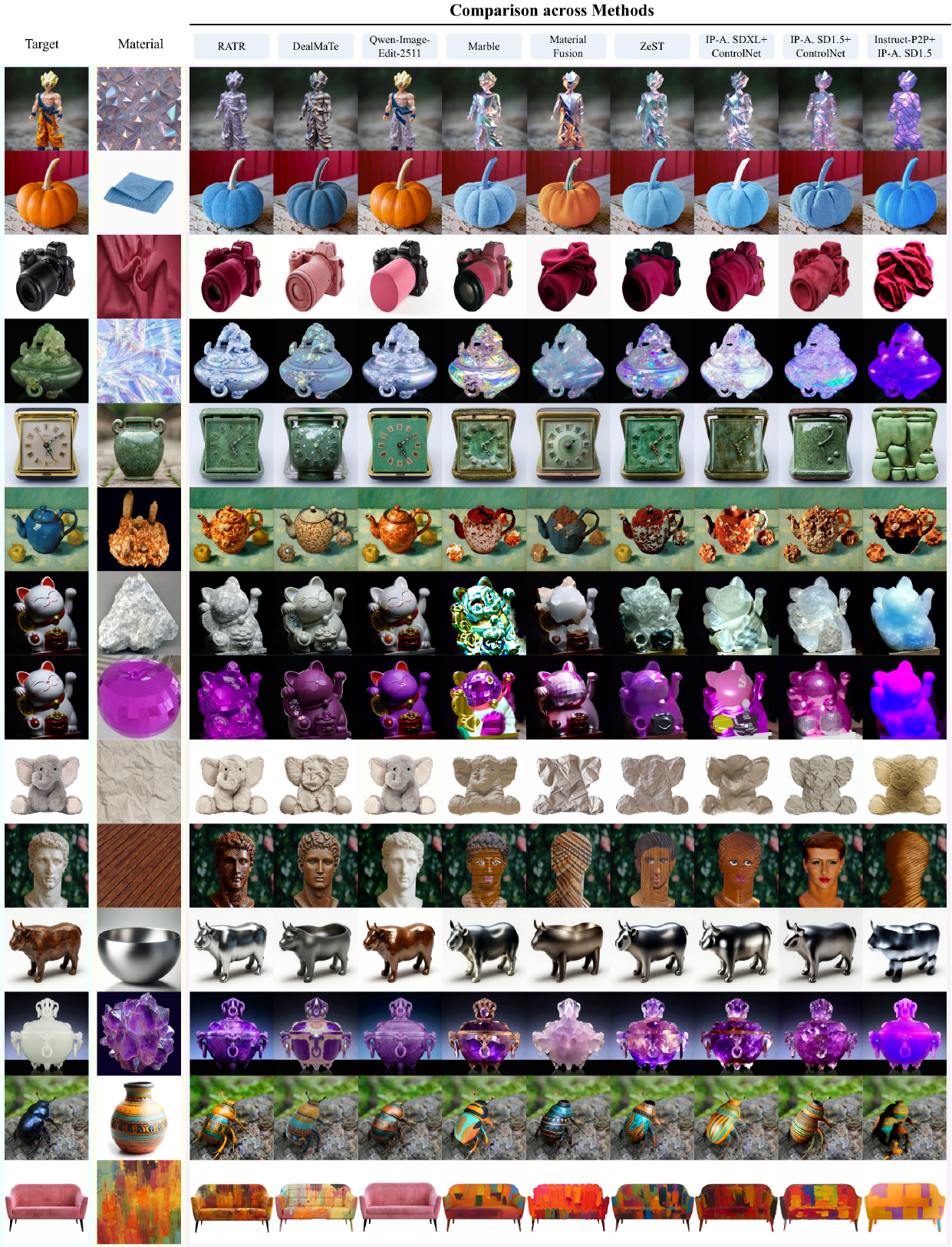}
 
\caption{
Qualitative comparison across methods. RATR achieves more faithful material transfer while better preserving target geometry, whereas competing methods often suffer from incomplete material transfer or structural artifacts. More qualitative results are provided in \href{https://tangjieh.github.io/RATR/}{Project}. }
\label{fig:qualitative_comparison}
\end{figure*}

\subsection{Inference Procedure}
\label{sec:inference}

At inference time, the target mask $M_s$ is computed once, and the reference-free release trajectory produces the early hypotheses $\{\widehat x_0(t):t\in\mathcal T_{\mathrm{rel}}\}$. Eq.~\ref{eq:release_selection} selects $t^\star$ separately for each target; luminance restoration then produces $C_s$. Independently, Material Evidence Isolation constructs $E_r$. The frozen editor encodes $(P,C_s,E_r)$ into $h$, $Z_s$, and $Z_r$. At each denoising step, SM-CDR applies the scheduled gains to the centered VAE residuals immediately before evaluating the denoiser, while retaining the original scheduler.

The editor's decoded result $\widetilde I_o$ follows the first visual input's generation canvas. After resizing it to the target resolution when necessary, the edited foreground is composited with the unchanged target background using the fixed mask:
\begin{equation}
I_o
=M_s\odot\widetilde I_o^{\uparrow}
+(1-M_s)\odot I_s,
\label{eq:final_compositing}
\end{equation}
where $\widetilde I_o^{\uparrow}$ is the decoded image aligned to the original target resolution. The entire procedure uses fixed model parameters and does not require material-specific training or pair-specific manual masks.

\section{Experiments}
\subsection{Setup}
\label{sec:exp_setup}

We evaluate RATR using 30 target images and 30 material exemplars
covering diverse object geometries and material appearances.
All methods are evaluated under the same protocol.
RATR uses Qwen-Image-Edit-2511~\citep{qwen2025imageedit2511}
as the frozen two-image editor with 30 denoising steps.
We compare with Qwen-Edit, the same frozen editor applied directly
to the raw target/reference pair, and representative transfer
baselines, including DealMaTe~\citep{huang2026dealmate},
MARBLE~\citep{cheng2025marble},
MaterialFusion~\citep{garifullin2025materialfusion},
ZeST~\citep{cheng2024zest},
IP-Adapter SDXL and IP-Adapter SD1.5 with
ControlNet~\citep{ye2023ipadapter,zhang2023controlnet},
and InstructPix2Pix with
IP-Adapter~\citep{brooks2023instructpix2pix,ye2023ipadapter}.

For all main experiments, we use a fixed three-stage routing
schedule with transitions at normalized denoising progress
$u_0=0.30$ and $u_1=0.70$.
The surface/material gains $(\alpha_s^\tau,\alpha_r^\tau)$
are set to $(1.20,0.80)$, $(0.85,1.35)$, and $(0.65,1.60)$
for the early, middle, and late stages, respectively,
yielding the material-routing-share profile
$\mathbf p_\Gamma=\{0.40,0.61,0.71\}$.
Surface release is parameterized by the fixed early candidate
set $\mathcal{T}_{\mathrm{rel}}=\{2,4,6,8,10\}$
on the 30-step trajectory.
Using normalized YCbCr, we set the neutral chroma anchor to
$c_0=(0.5,0.5)$ and use fixed release-energy weights
$\lambda_{\mathrm{alb}}=1.0$,
$\lambda_{\mathrm{geo}}=1.2$, and
$\lambda_{\mathrm{sup}}=0.8$.
For each target image, the release step is automatically selected
according to Eq.~\ref{eq:release_selection} by minimizing
the three-term release energy over $\mathcal{T}_{\mathrm{rel}}$.
The resulting surface condition is reused across different
material references for the same target.
We preserve the input aspect ratio and limit the longer
side to 1024 pixels.
All main results use the same predefined hyperparameters
and automatic release-selection procedure, without
manual per-example tuning.
RATR and the backbone-matched Qwen-Edit baseline use
the same material-rebinding instruction across all
target--material pairs.

\begin{table*}[t]
\caption{Target-side quantitative comparison results of our method and the existing competitors. Higher is better for both SSIM and Edge SSIM; lower is better for LPIPS, DreamSim, and Edge MAE. The best results are marked in \textbf{bold}. }
\label{tab:quantitative_comparison}
\centering
\small
\begin{tabular*}{\textwidth}{@{\extracolsep{\fill}}lccccc@{}} 
\toprule
Method & SSIM $\uparrow$ & LPIPS $\downarrow$ & DreamSim $\downarrow$ & Edge SSIM $\uparrow$ & Edge MAE $\downarrow$ \\
\midrule
DealMaTe & 0.7061 & 0.3927 & 0.3887 & 0.7190 & 0.0158 \\
Qwen-Edit & 0.7051 & 0.3560 & 0.3430 & 0.6948 & 0.0163 \\
MARBLE & 0.6356 & 0.3957 & 0.4264 & 0.6464 & 0.0204 \\
MaterialFusion & 0.7178 & 0.3419 & 0.3886 & 0.6950 & 0.0162 \\
ZeST & 0.6648 & 0.3781 & 0.4314 & 0.6722 & 0.0186 \\
IP-A. SDXL+CN & 0.6152 & 0.4142 & 0.4795 & 0.6463 & 0.0204 \\
IP-A. SD1.5+CN & 0.5978 & 0.4407 & 0.5385 & 0.6191 & 0.0228 \\
Inst-P2P+IP-A. & 0.6906 & 0.3928 & 0.5144 & 0.7042 & 0.0163 \\
\midrule
\FullRowBackground{aliceblue!60} RATR (Ours) & \textbf{0.7292} & \textbf{0.3127} & \textbf{0.3414} & \textbf{0.7282} & \textbf{0.0153} \\
\bottomrule
\end{tabular*}
\end{table*}

\begin{table}[t]
\caption{
Reference-material correspondence comparison.
Higher is better for all metrics.
}
\label{tab:material}
\centering
\begin{tabular}{lccc}
\toprule
Method
& CLIP $\uparrow$
& DINOv3 $\uparrow$
& Materialistic $\uparrow$ \\

\midrule
DealMaTe
& 0.4617
& 0.5019
& 0.5974 \\

Qwen-Edit
& 0.4928
& 0.5322
& 0.5323 \\

MARBLE
& 0.5058
& 0.5249
& 0.5643 \\

MaterialFusion
& 0.4810
& 0.5313
& 0.4594 \\

ZeST
& 0.5247
& 0.5456
& 0.5817 \\

IP-A-SDXL-CN
& 0.5249
& 0.5437
& 0.6060 \\

IP-A-SD15-CN
& 0.5246
& 0.5470
& 0.6217 \\

InstP2P-IP-A
& 0.5175
& 0.5351
& 0.5958 \\

\midrule
\rowcolor{aliceblue!60} RATR (Ours)
& \textbf{0.5251}
& \textbf{0.5495}
& \textbf{0.6225} \\
\bottomrule
\end{tabular}
\end{table}

\subsection{Main Results}
\label{sec:main_results}

\subsubsection{Qualitative Comparison}

Fig.~\ref{fig:qualitative_comparison} compares RATR with Qwen-Edit
and representative material-transfer methods across diverse
target--material pairs. One recurring failure of existing methods
is incomplete replacement of the target's original appearance.
For example, Qwen-Edit largely retains the pumpkin's orange
appearance despite the blue fabric reference, leaves the marble
bust nearly unchanged with the wood reference (the tenth row
of Fig.~\ref{fig:qualitative_comparison}), and preserves the
cow's brown appearance despite the metallic exemplar
(the fourth-to-last row of Fig.~\ref{fig:qualitative_comparison}).
In contrast, RATR gives the pumpkin a blue fabric-like appearance
while preserving its grooves and stem, imparts wood grain to the
bust while retaining its recognizable facial structure, and
reproduces metallic reflections on the cow while preserving
its horns and legs.

Other competing methods produce more pronounced appearance
changes but may introduce reference-specific structures or
distort target details. In the fifth row of
Fig.~\ref{fig:qualitative_comparison}, DealMaTe introduces
vase-like handles from the green ceramic reference, whereas
RATR adopts the reference material appearance while preserving
the clock's rectangular frame and face. For the camera with a
red fabric reference (the third row of
Fig.~\ref{fig:qualitative_comparison}), MaterialFusion transforms
the lens and body into folded fabric-like forms, while RATR
maintains the recognizable lens barrel and camera silhouette.
Similar differences appear in the lucky cat with a rock
reference and the plush elephant with a crumpled-paper
reference. Some methods distort the cat's raised paw and
facial features or weaken the elephant's distinctive ears
and trunk. In contrast, RATR better preserves these
recognizable structures while adapting their surface
appearances to the corresponding materials.

\begin{figure*}[!tp]
\centering
\includegraphics[width=\textwidth]{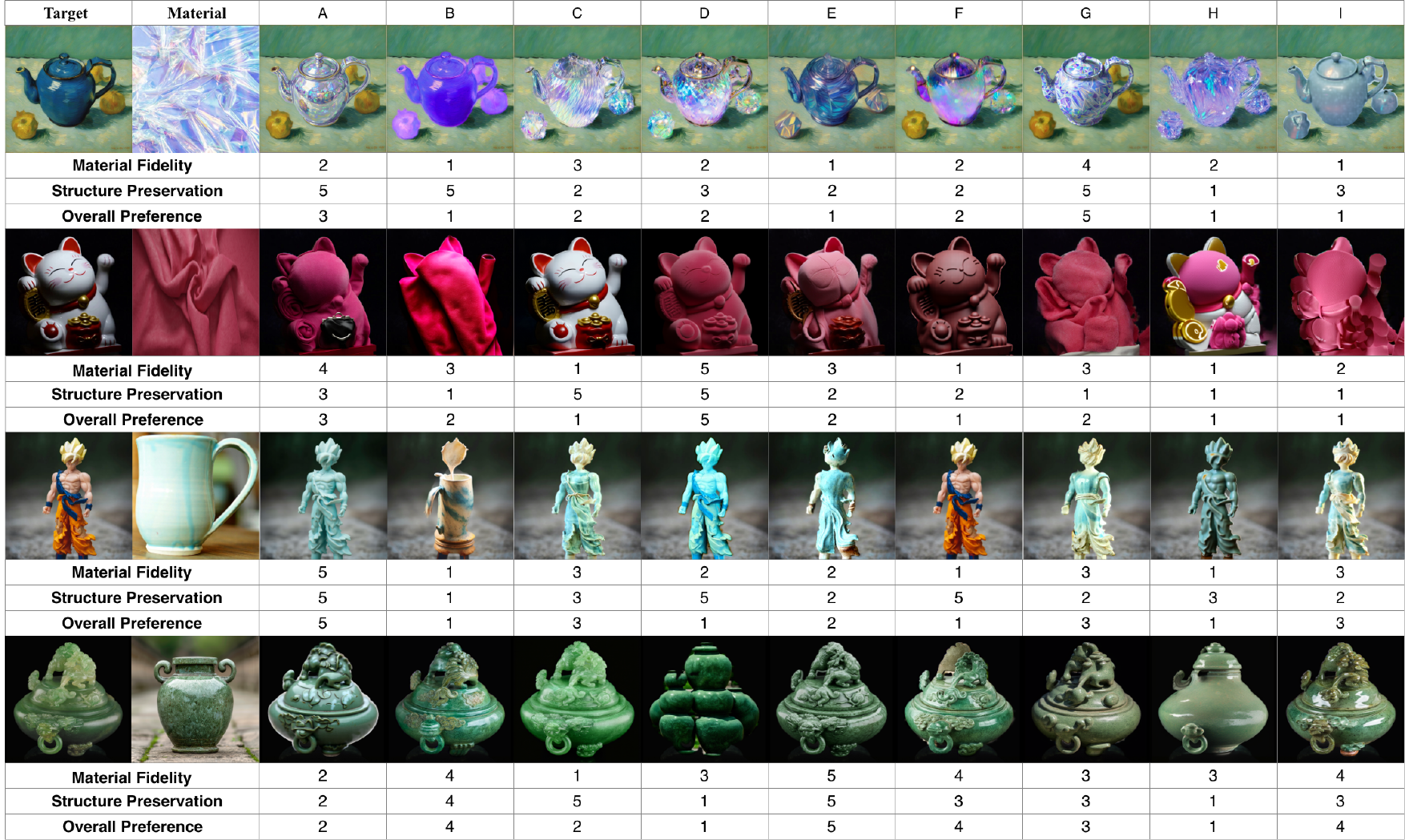}
\vspace{0.01cm}
\caption{
Representative human ratings across four target--material pairs.
Each row presents the target, material reference, and nine
anonymized outputs (A--I), together with their five-point
Material Fidelity, Structure Preservation, and Overall Preference
ratings. RATR corresponds to G, D, A, and E from top to bottom.
The displayed scores illustrate individual cases, while
aggregate evaluation results are reported in
Table~\ref{tab:human_evaluation}.
}
\label{fig:human_eval_examples}
\end{figure*}

\begin{figure*}[t]
    \centering
    \begin{overpic}[width=0.95\linewidth]{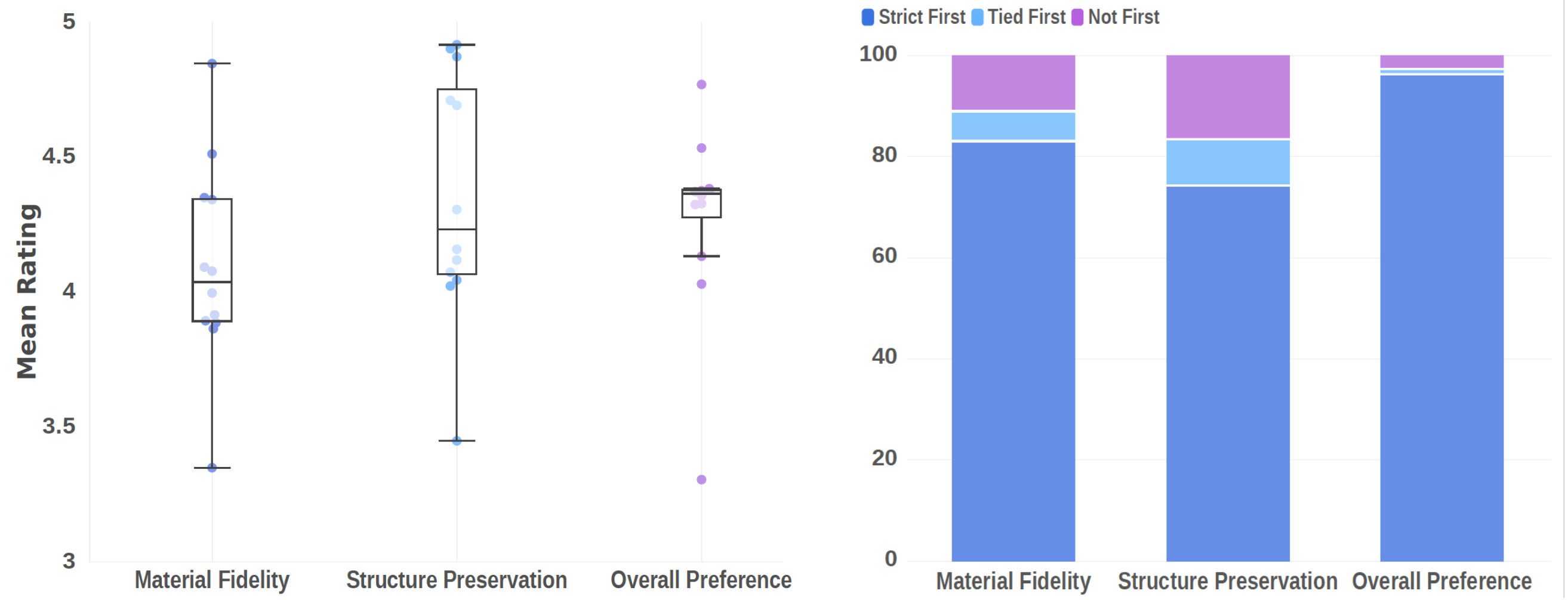}
        \put(15, -3){{\scriptsize (a) Evaluator ratings}}
        \put(70, -3){{\scriptsize (b) First-place rates}}
    \end{overpic}
    \vspace{0.5cm}
  \caption{
Human-rating distributions and pair-level rankings for RATR.
(a) Evaluator-level mean ratings across the three criteria;
dots represent individual evaluators, boxes indicate
interquartile ranges, and horizontal lines denote medians.
(b) Proportions of target--material pairs where RATR
ranks strictly first, ties for first, or is not first
among the nine methods.
}
\label{fig:human_eval_distribution}
\end{figure*}

\subsubsection{Quantitative Comparison}

Table~\ref{tab:quantitative_comparison} reports SSIM,
LPIPS, DreamSim, Edge SSIM, and Edge MAE.
SSIM, LPIPS, and DreamSim assess target-side structural
and perceptual consistency, while both Edge SSIM and
Edge MAE provide edge-based diagnostics of silhouette
and boundary preservation.

RATR achieves the best aggregate results on all five
reported target-side preservation metrics in
Table~\ref{tab:quantitative_comparison}.
Its SSIM, LPIPS, and DreamSim results indicate greater
structural and perceptual consistency with the target,
while Edge SSIM and Edge MAE show better preservation
of silhouettes and boundaries.
Table~\ref{tab:material} further shows that RATR
achieves the highest mean scores on all three
reference-side metrics, including CLIP, DINOv3,
and Materialistic.
Together, these results demonstrate that improved
target preservation is accompanied by leading
reference-material correspondence, rather than
a reduction in material alignment.
This joint performance supports the qualitative
observation that RATR mitigates the trade-off
between target preservation and material transfer.
The observed balance is consistent with its
role-assignment and temporal-routing design,
which reduces interference from the original target
appearance while coordinating
structural and material evidence during denoising.

\begin{table*}[t]
\caption{
Human evaluation of material fidelity, target-structure preservation, and overall preference. Scores are reported as the mean $\pm$ standard deviation of evaluator-level mean ratings on a 1--5 scale; higher scores indicate better performance.
}
\label{tab:human_evaluation}
\centering
\scriptsize
\begin{tabular*}{\textwidth}{@{\extracolsep{\fill}}lccc@{}} 
\toprule
Method & Material Fidelity $\uparrow$ & Structure Preservation $\uparrow$ & Overall Preference $\uparrow$ \\
\midrule
DealMaTe & 3.009 $\pm$ 0.138 & 3.932 $\pm$ 0.622 & 3.165 $\pm$ 0.234 \\
Qwen-Edit & 3.239 $\pm$ 0.229 & 3.792 $\pm$ 0.512 & 3.328 $\pm$ 0.176 \\
MARBLE & 3.221 $\pm$ 0.207 & 3.381 $\pm$ 0.415 & 3.137 $\pm$ 0.205 \\
MaterialFusion & 2.607 $\pm$ 0.337 & 3.068 $\pm$ 0.457 & 2.626 $\pm$ 0.344 \\
ZeST & 3.203 $\pm$ 0.341 & 3.158 $\pm$ 0.404 & 3.016 $\pm$ 0.285 \\
IP-A. SDXL+CN & 3.222 $\pm$ 0.312 & 2.581 $\pm$ 0.753 & 2.531 $\pm$ 0.573 \\
IP-A. SD1.5+CN & 3.325 $\pm$ 0.455 & 2.361 $\pm$ 0.786 & 2.488 $\pm$ 0.575 \\
Inst-P2P+IP-A. & 2.440 $\pm$ 0.430 & 2.450 $\pm$ 0.436 & 2.070 $\pm$ 0.402 \\
\midrule
\FullRowBackground{aliceblue!60} RATR (Ours) & \textbf{4.093} $\pm$ 0.381 & \textbf{4.355} $\pm$ 0.460 & \textbf{4.273} $\pm$ 0.355 \\
\bottomrule
\end{tabular*}
\end{table*}

\begin{table*}[!t]
\centering
\caption{Quantitative comparison of SM-CDR routing strategies.}
\label{tab:ablation_smcdr}
\footnotesize
\setlength{\tabcolsep}{3.5pt}
\begin{tabular*}{\textwidth}{@{\extracolsep{\fill}}lcccccccc@{}} 
\toprule
Routing Strategy
& SSIM $\uparrow$
& LPIPS $\downarrow$
& DreamSim $\downarrow$
& Edge SSIM $\uparrow$
& Edge MAE $\downarrow$
& CLIP $\uparrow$
& DINOv3 $\uparrow$
& Materialistic $\uparrow$ \\
\midrule
Structure-lock
& \textbf{0.7502}
& \textbf{0.2949}
& \textbf{0.2910}
& \textbf{0.7476}
& \textbf{0.0135}
& 0.4722
& 0.5140
& 0.5062 \\

Default
& 0.7292
& 0.3127
& 0.3414
& 0.7282
& 0.0153
& 0.5251
& 0.5495
& 0.6225 \\

Material-forward
& 0.7098
& 0.3326
& 0.3687
& 0.7087
& 0.0171
& \textbf{0.5390}
& \textbf{0.5633}
& \textbf{0.6531} \\
\bottomrule
\end{tabular*}
\end{table*}

\subsubsection{Human Evaluation}

To complement the automated similarity metrics, we conduct a blind
human evaluation along three perceptual dimensions:
\emph{Material Fidelity}, \emph{Structure Preservation}, and
\emph{Overall Preference}. Material Fidelity measures how closely an output reproduces
the reference material's texture, color variation, gloss or roughness,
surface detail, and natural coverage.
For example, transferring blue fabric to a pumpkin should reproduce
fabric-like surface qualities, not merely recolor the pumpkin blue. Structure Preservation evaluates retention of the target's
identity, silhouette, geometry, pose, major components,
relative layout, and boundaries.
For a clock conditioned on a green ceramic vase, the rectangular
frame and clock face should remain intact, while transferring
the vase's handles would constitute a structural error. Overall Preference assesses the complete rebinding result
by jointly considering material appearance, structural integrity,
visual naturalness, and artifacts.

12 graduate students with research backgrounds in computer vision
and graphics participated as independent evaluators.
The evaluation covers 272 target--material pairs and nine
compared methods.
For each pair, evaluators view the target image,
the material reference, and anonymized, randomly ordered method outputs
under the same interface.
Each output is assessed separately on all three criteria using an
integer scale from 1 to 5, where 1 denotes a clear failure,
3 indicates partial success with noticeable defects,
and 5 denotes a very high degree of satisfaction;
2 and 4 represent intermediate levels.
Different methods may receive the same rating.
The protocol allows up to 88,128 criterion-level ratings,
and the reported means are computed from valid ratings
for each method and criterion.

\begin{figure}
\centering
\includegraphics[width=0.99\columnwidth]{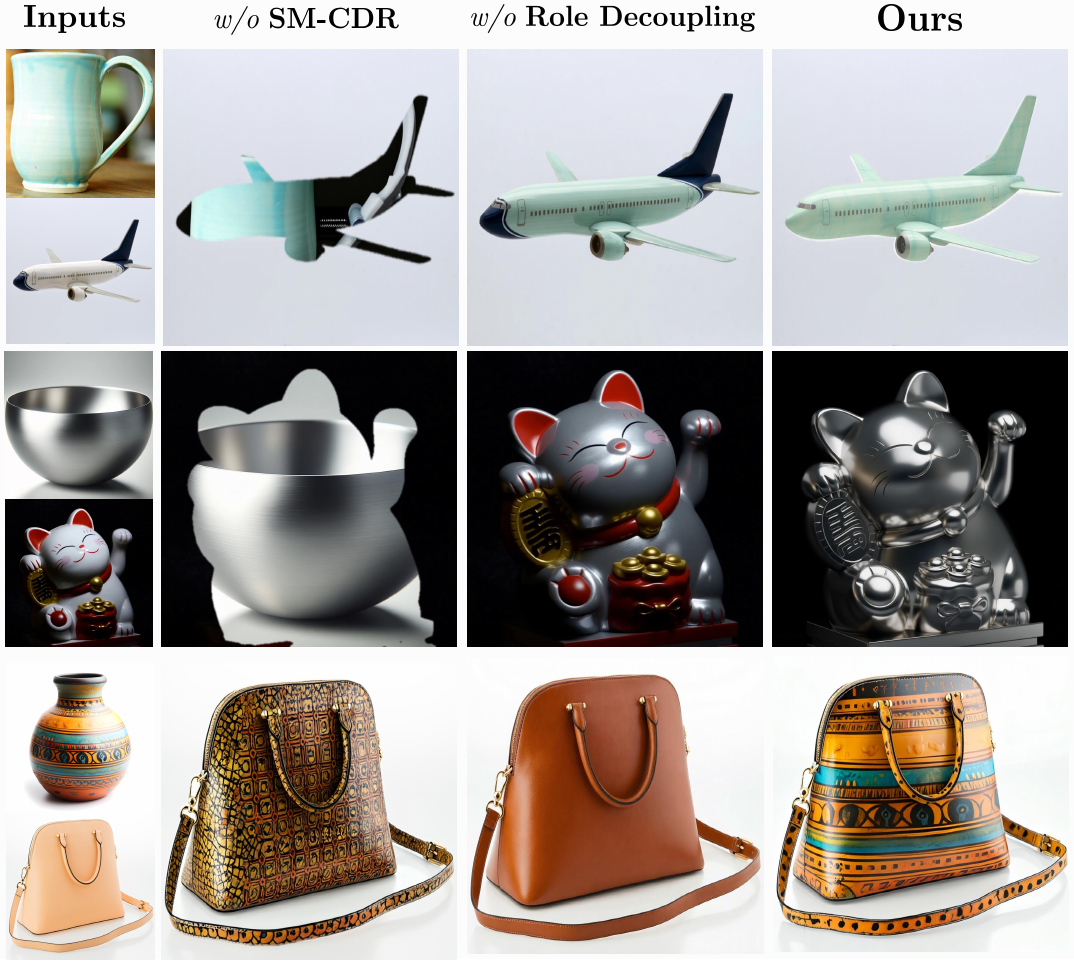}
\caption{Ablation of target-side role decoupling and SM-CDR.}
\label{fig:ablation_components}
\end{figure}


Fig.~\ref{fig:human_eval_examples} provides representative examples
of how material fidelity and structural preservation are jointly
assessed across different target geometries and material appearances.
The iridescent teapot and fabric-conditioned lucky cat illustrate
the importance of reproducing distinctive surface characteristics
while retaining recognizable components such as the teapot's
spout and handle or the cat's raised paw and facial features.
The ceramic-conditioned figurine and ornamental vessel further
demonstrate the need to preserve target-specific geometry,
pose, and surface relief without introducing structural
characteristics from the material reference.

\begin{figure*}[!tp]
\centering
\includegraphics[width=0.99\textwidth]{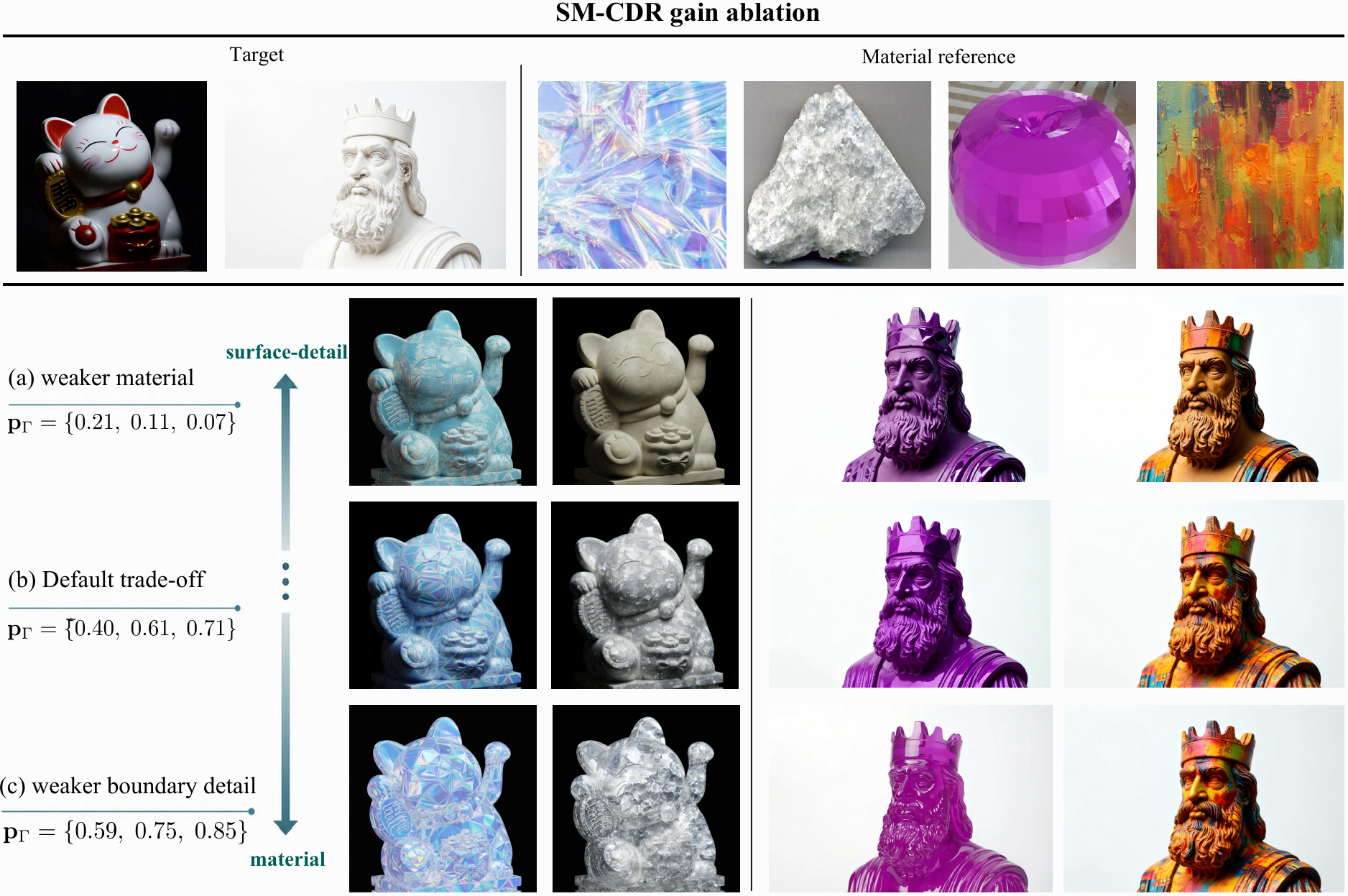}
\caption{SM-CDR gain ablation under different
material-routing-share profiles $\mathbf{p}_\Gamma$.}
\label{fig:ablation_smcdr_gain}
\end{figure*}

Table~\ref{tab:human_evaluation} presents the human evaluation results of 12 evaluators. From the results, we observe that RATR achieves the highest mean rating on all three criteria, exceeding the strongest competitor in each criterion by 0.768, 0.423, and 0.945 points, respectively. Its simultaneous advantage in material fidelity and structure preservation indicates that improved reference-material adoption is accompanied by better retention of target identity and geometry. The highest Overall Preference rating further suggests that this balance translates into higher perceived quality of the complete rebinding results.

Beyond aggregate comparisons, Fig.~\ref{fig:human_eval_distribution}
provides complementary analyses of RATR's performance across
evaluators and target--material pairs.
Fig.~\ref{fig:human_eval_distribution}(a) shows the distributions
of evaluator-level mean ratings for the three perceptual criteria.
For both Structure Preservation and Overall Preference,
11 of the 12 evaluators assign RATR mean ratings of at least 4.0.
These distributions complement the aggregate scores in
Table~\ref{tab:human_evaluation}, showing that the favorable
structural and overall assessments extend across the evaluator
group rather than being concentrated in a few individual judgments.

Fig.~\ref{fig:human_eval_distribution}(b) further examines
the prevalence of RATR's advantage across different
target--material combinations.
Based on the mean ratings for each pair and criterion,
RATR ranks strictly first among all nine methods on
226 of 272 pairs (83.1\%) in Material Fidelity,
202 pairs (74.3\%) in Structure Preservation,
and 262 pairs (96.3\%) in Overall Preference.
In particular, the exceptionally high first-place rate
in Overall Preference indicates that RATR's perceptual
advantage extends to nearly all evaluated combinations.
These results demonstrate that its leading mean scores
are accompanied by broad pair-level superiority,
rather than being driven by a limited subset of
particularly favorable examples.

The representative scoring,
aggregate comparisons, evaluator-level distributions,
and pair-level rankings provide complementary
perceptual evidence of RATR's effectiveness.
The consistent advantages in material fidelity
and structural preservation support the central
objective of role-consistent material rebinding:
transferring reference appearance while retaining
target-specific geometry and identity.
These observations are consistent with the coordinated
role assignment and temporal routing formulation,
whose individual contributions are further examined
in the ablation study.

\begin{figure*}[!tp]
\centering
\includegraphics[width=0.99\linewidth]{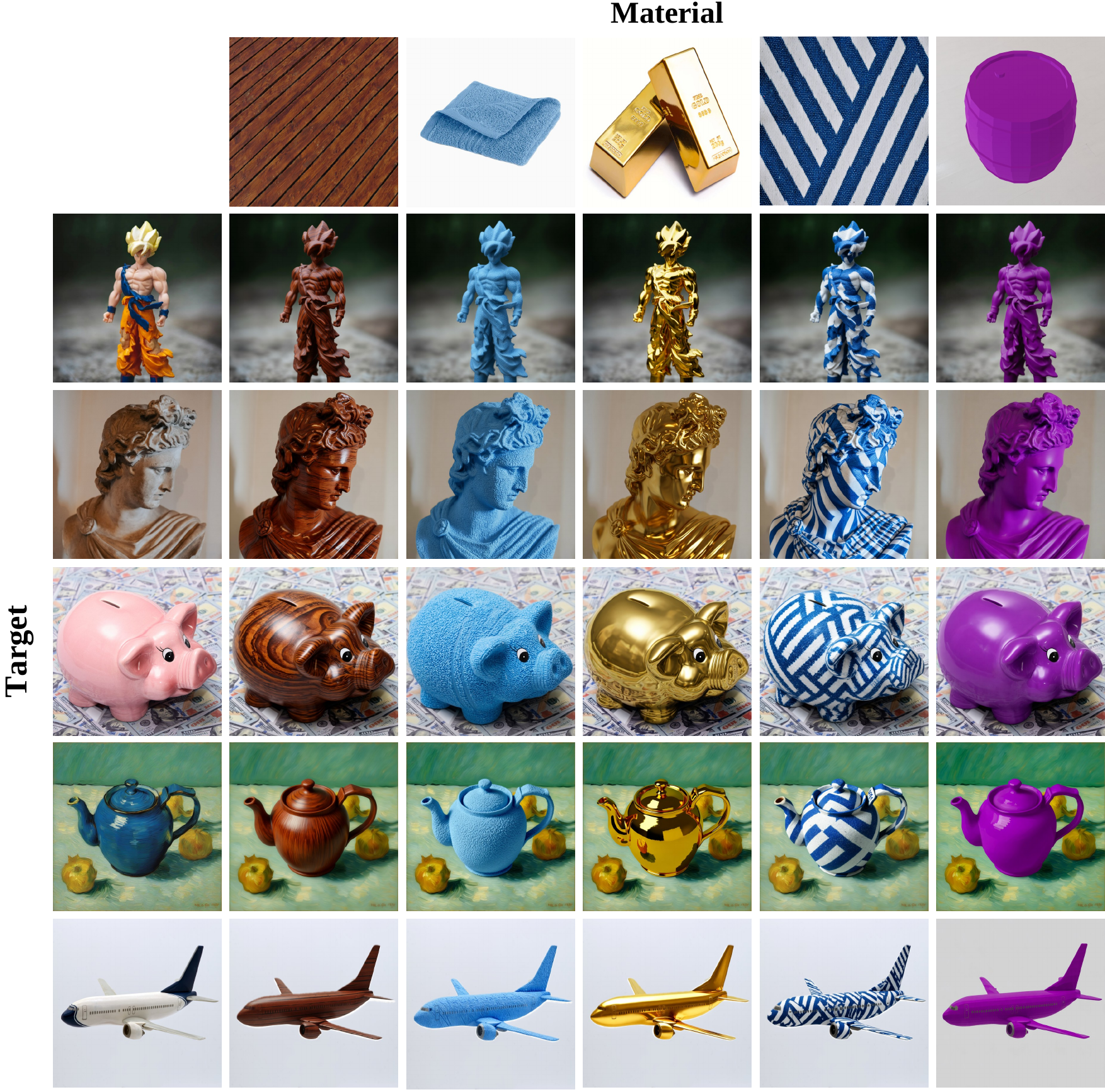}
\caption{
Additional qualitative results across diverse target geometries and
material appearances.
}
\label{fig:additional_visualization_1}
\end{figure*}

\begin{figure*}[!tp]
\centering
\includegraphics[width=0.99\linewidth]{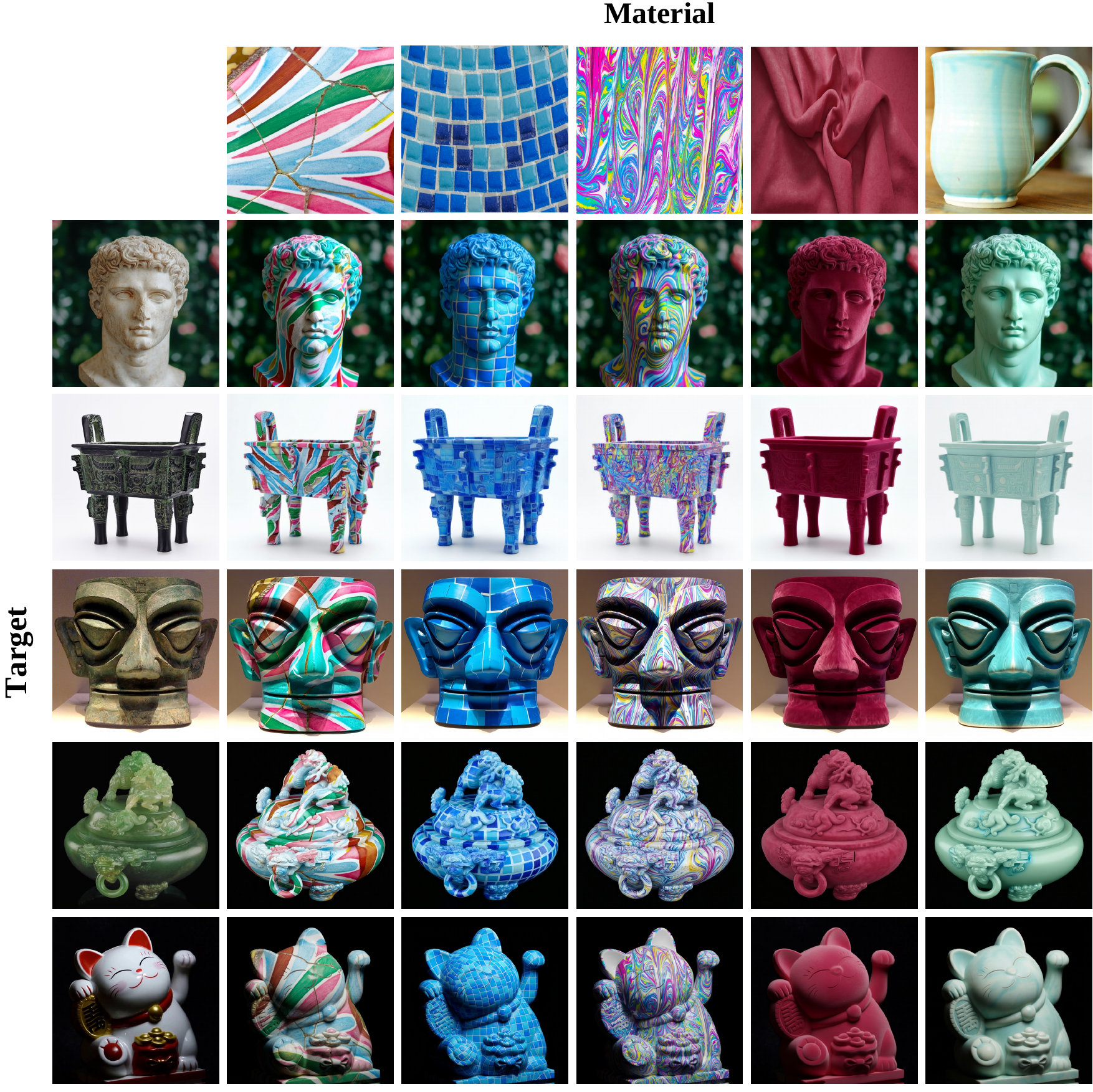}
\caption{
Additional qualitative results with patterned and object-centric
material references.
}
\label{fig:additional_visualization_2}
\end{figure*}

\begin{figure*}[!tp]
\centering
\includegraphics[width=0.99\textwidth]{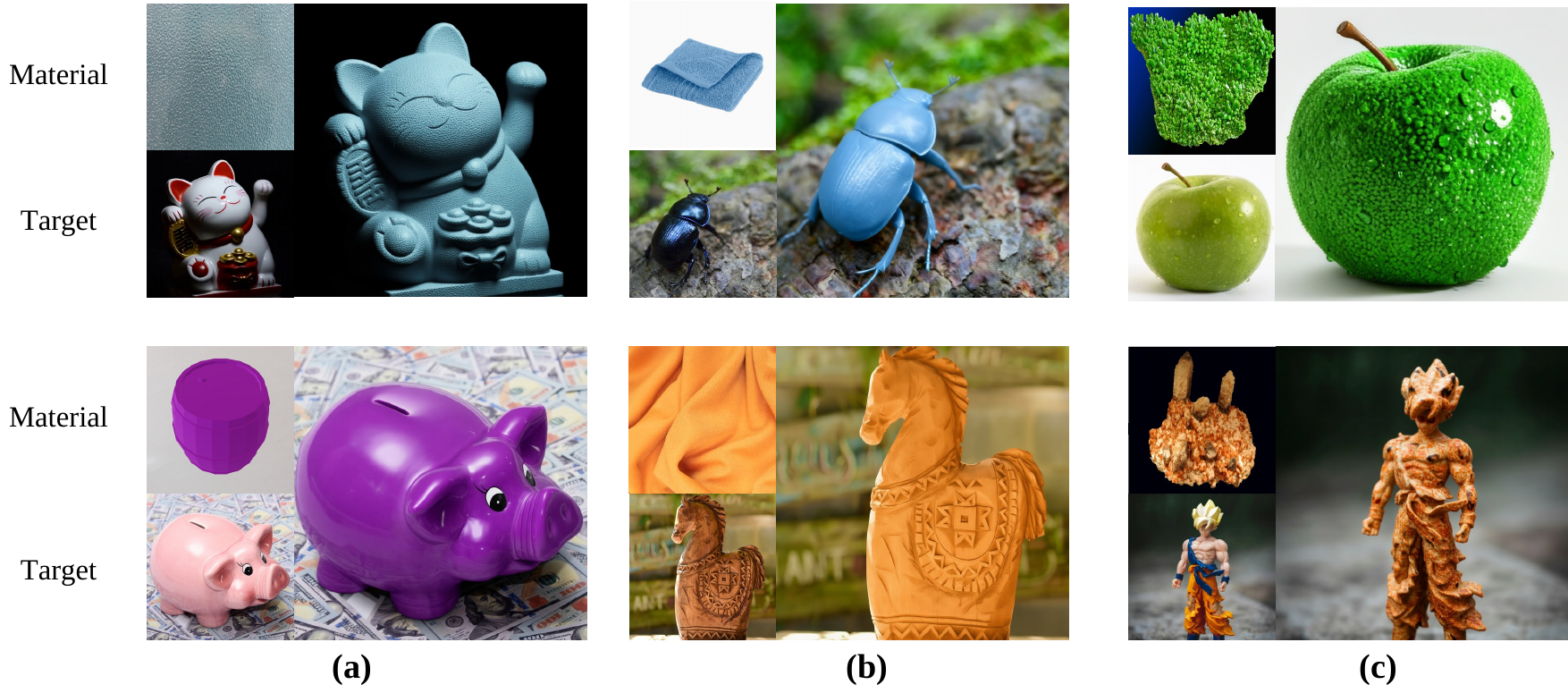}
\caption{
Representative limitations of RATR.
(a) Residual dependence of material reflectance on target illumination.
(b) Pair-dependent material adoption under shared routing.
(c) Limited reproduction of material-specific geometric relief
under target-structure preservation.
}
\label{fig:limitations}
\end{figure*}

\subsection{Ablation Study}
\label{sec:ablation}

In this section, we examine the contributions of
role decoupling and SM-CDR, followed by an analysis
of different SM-CDR routing strategies.

Fig.~\ref{fig:ablation_components} examines the two
central mechanisms of RATR: role decoupling and
temporal routing. Removing role decoupling restores
the raw target image in place of $C_s$, while retaining
the reference-side material evidence $E_r$ and SM-CDR.
This reintroduces the coupling between target geometry
and its original appearance, weakening the intended
complementarity of the two visual conditions.
In contrast, removing SM-CDR preserves the role-specific
visual conditions but returns to static use of their
VAE condition segments without temporal modulation.

The visual differences reveal the complementary effects
of these two mechanisms. Without role decoupling,
the raw target condition carries both desired geometry
and undesired source appearance, making the output
more susceptible to original color, gloss, or texture
retention. Without SM-CDR, the role-specific conditions
lack temporal coordination, leading to less coherent
material adoption and structural preservation in the
shown examples. The full model combines role-consistent
condition construction with temporal routing, yielding
clearer material expression while retaining the
target-specific structures.

Fig.~\ref{fig:ablation_smcdr_gain} analyzes the
controllability of SM-CDR by varying the
material-routing-share profile $\mathbf{p}_\Gamma$.
A structure-dominant route with a smaller material
share preserves more surface detail and boundary
support but under-transfers the reference appearance,
whereas a material-forward route strengthens the
transferred material at the cost of weaker boundary
detail and local structure. The default route,
$\mathbf{p}_\Gamma=\{0.40,0.61,0.71\}$, provides a
strong visual balance among the shown settings by
retaining structural support early and increasing
material influence later.

Table~\ref{tab:ablation_smcdr} further reveals how routing emphasis affects the balance between structural preservation and material transfer. Structure-lock performs best on all five target-preservation metrics but yields weaker material correspondence. This suggests that strong structural conditioning constrains appearance changes as well as geometric deviations, preserving the target more closely while limiting reference-material adoption. Conversely, Material-forward achieves the highest material-side scores at the expense of structural consistency, indicating that stronger material expression can also alter target boundaries and fine details.

The default strategy balances these competing effects. Although it does not maximize either group of metrics, it preserves structure more effectively than Material-forward and transfers material more faithfully than Structure-lock. Importantly, these results show that role-specific condition construction does not eliminate the need to regulate their influence: even after reparameterization, excessive emphasis on either condition can undermine the complementary objective. Consistent with the visual comparisons in Fig.~\ref{fig:ablation_smcdr_gain}, the default schedule provides a practical balance between structural constraint and appearance adaptation, while the alternative routes demonstrate the controllability of this trade-off.


\subsection{Additional Qualitative Visualizations}
\label{sec:additional_visualizations}

Figs.~\ref{fig:additional_visualization_1} and
\ref{fig:additional_visualization_2} provide additional qualitative
results under complementary material-rebinding settings. In each grid,
material references appear in the top row, target objects in the left
column, and the remaining cells show the corresponding rebinding results.

Fig.~\ref{fig:additional_visualization_1} emphasizes diversity in both
target geometry and material appearance, covering substantially different
object categories together with materials ranging from wood and fabric to
metallic, patterned, and glossy surfaces. Across these combinations, RATR
preserves target-specific geometry and local structural characteristics
while adapting the visible surface appearance to the reference material.

Fig.~\ref{fig:additional_visualization_2} further considers patterned and
object-centric material references, where the reference may contain stronger
spatial patterns or reference-specific structural cues. The results show
that the same material appearance can be transferred across targets with
different geometries while maintaining the target structure and limiting
visible reference-object interference.

\FloatBarrier

\section{Scope and Limitations}
\label{sec:limitations}

RATR is designed as an inference-time material rebinding method
for a frozen two-image editor. Its goal is to reorganize the
target and reference visual conditions into a more role-consistent
form, rather than to train a material-specific generator or
recover physically based material parameters. Consequently,
its performance is bounded by the material and editing priors
already available in the frozen editor. For materials that require
highly precise physical effects, such as complex refraction,
mirror-like multi-bounce reflection, or very fine anisotropic
microstructure, RATR can improve the organization and timing
of visual evidence, but it does not guarantee physically exact
material synthesis. As illustrated in
Fig.~\ref{fig:limitations}(a), the reference color and surface
characteristics can be transferred while the resulting appearance
remains partly influenced by the target's original shading.
In particular, fine-scale specular highlights and material-specific
reflection patterns may not be faithfully reproduced, as RATR
does not explicitly recover material reflectance or scene illumination.

SM-CDR provides a controllable routing interface
rather than a global optimization guarantee. The material
routing share can adjust the balance between material
expression and surface-detail preservation, but a single
fixed policy cannot be optimal for every target--material
pair. Fig.~\ref{fig:limitations}(b) illustrates this
variation, where the reference colors are successfully
adopted while their characteristic fine-scale textures
remain less pronounced than the target's original surface
details. Although the default routing policy provides
a favorable overall balance, its effects can vary with
the geometry and appearance characteristics of individual
target--material combinations. Very material-forward routes
may weaken boundary details, while overly surface-dominant
routes may under-transfer the reference material.


The albedo-resolved surface boundary is also a role-specific
condition, not a full intrinsic decomposition. It suppresses
target-side chromatic appearance while preserving geometry-dominant
luminance structure. This design is effective for reducing
old-color retention, but it may be less suitable when the target
identity is itself defined by non-geometric color markings,
printed text, or semantic logos. In such cases, aggressively
weakening source chroma may remove information that is visually
important to the object. RATR mitigates this by restoring
luminance-side structural residuals, but it does not explicitly
reason about semantic markings. Moreover, preserving the target
geometry can limit the reproduction of materials with pronounced
three-dimensional surface relief. As shown in
Fig.~\ref{fig:limitations}(c), the characteristic color and
surface appearance of the reference are transferred, while
prominent protrusions and angular material structures are
less fully reproduced. This reflects the intended scope of
RATR as a geometry-preserving appearance-rebinding method
rather than a material-driven geometry reconstruction approach.


\section{Conclusion}
\label{sec:conclusion}

We presented RATR, a training-free material rebinding framework
that addresses the coupled structural and appearance information
in reference-conditioned image editing through role assignment
and temporal routing. By constructing complementary
surface-oriented and material-oriented visual conditions,
RATR reduces interference from the original target appearance
and reference-specific content. SM-CDR further coordinates
their centered VAE condition residuals throughout denoising,
allowing structural preservation and material adoption
to be balanced within a frozen two-image editor.

Extensive evaluations demonstrate improved target-side
structural preservation alongside leading reference-side
material fidelity, with human assessments further supporting
the perceptual quality of the resulting material rebinding.
The ablation results highlight the complementary roles
of condition construction and temporal coordination.
Together, these findings suggest that effective material
rebinding depends not only on the visual evidence provided
to a generative editor, but also on how that evidence
is assigned and coordinated during synthesis.

\section*{Statements and Declarations}

\noindent\textbf{Funding.}
This research was supported in part by the National Natural Science Foundation of China under Grant 62576313, the Key R\&D  Program of Zhejiang Province under Grant 2026C01016, and the Key R\&D  Program of Ningbo under Grant 2026T018.


\noindent\textbf{Competing interests.}
The authors declare that they have no relevant financial or
non-financial interests to disclose.

\noindent\textbf{Data availability.}
The data supporting the findings of this study are available
from the corresponding author upon reasonable request.

\noindent\textbf{Code availability.}
The implementation of RATR is available from the corresponding
author upon reasonable request.

\bibliography{main}

\end{document}